\documentclass[letterpaper, 10 pt, conference]{ieeeconf}  

\IEEEoverridecommandlockouts                              

\usepackage{amsmath} 
\usepackage{amssymb}  

\usepackage{graphicx}
\usepackage{capt-of} 
\usepackage{overpic} 
\usepackage[export]{adjustbox} 
\usepackage{textcomp} 
\usepackage{tabularx}
\usepackage{multirow}
\usepackage{booktabs}
\usepackage{arydshln}
\usepackage{adjustbox} 
\usepackage{pgfplots} 
\usepackage{units}
\usepackage{siunitx}
\usepackage[implicit=false]{hyperref}

\usepackage[caption=false, font=scriptsize]{subfig} 

\newcommand\copyrighttext{\footnotesize \textcopyright~2022 IEEE. Personal use of this material is permitted.  Permission from IEEE must be obtained for all other uses, in any current or future media, including reprinting/republishing this material for advertising or promotional purposes, creating new collective works, for resale or redistribution to servers or lists, or reuse of any copyrighted component of this work in other works.
}%

\newcommand\copyrightnotice{%
	\begin{tikzpicture}[remember picture,overlay]
	\node[anchor=south,xshift=0pt,yshift=14pt] at (current page.south) {\fbox{\parbox{\dimexpr\textwidth-\fboxsep-\fboxrule\relax}{\copyrighttext}}};
	\end{tikzpicture}%
}

\title{\LARGE \bf
MEAT: Maneuver Extraction from Agent Trajectories
}

\author{Julian Schmidt$^{1, 2}$, Julian Jordan$^{1}$, David Raba$^{1}$, Tobias Welz$^{1}$ and Klaus Dietmayer$^{2}$
\thanks{$^{1}$Mercedes-Benz AG, R\&D, Stuttgart, Germany \newline {\tt\small julian.sj.schmidt@mercedes-benz.com }
        }%
\thanks{$^{2}$Ulm University, Institute of Measurement, Control and Microtechnology, Ulm, Germany}%
}

\begin{document}

\maketitle
\thispagestyle{empty}
\pagestyle{empty}

\begin{abstract}
Advances in learning-based trajectory prediction are enabled by large-scale datasets.
However, in-depth analysis of such datasets is limited.
Moreover, the evaluation of prediction models is limited to metrics averaged over all samples in the dataset.
We propose an automated methodology that allows to extract maneuvers (e.g., left turn, lane change) from agent trajectories in such datasets.
The methodology considers information about the agent dynamics and information about the lane segments the agent traveled along.
Although it is possible to use the resulting maneuvers for training classification networks, we exemplary use them for extensive trajectory dataset analysis and maneuver-specific evaluation of multiple state-of-the-art trajectory prediction models.
Additionally, an analysis of the datasets and an evaluation of the prediction models based on the agent dynamics is provided.

\end{abstract}

\section{Introduction}
\copyrightnotice 
Learning-based trajectory prediction is one of the most competitive research fields related to motion planning of autonomous vehicles.
There are two main reasons that contribute to this:
Firstly, datasets required for training are obtainable without manual label effort.
Creating such a dataset requires to split recordings of a perception and tracking stack into two parts.
The first part of the recording forms the input data, while the second part of the recording is used as ground-truth.
Using this approach, several large-scale datasets (e.g., \cite{Chang2019, Sun2020}) have already been created and made publicly available.
Secondly, given these large-scale datasets, models with an arbitrary amount of complexity can be trained end-to-end in a supervised manner.

While many researchers use these datasets and try to top the benchmarks on one of the standard metrics, one thing falls by the wayside:
An in-depth analysis of the used datasets.
Such an in-depth analysis does not only help to understand dataset balance, but also enables a detailed evaluation of prediction models, which should not be limited to a few standard metrics averaged over the whole dataset.
This detailed evaluation makes it possible to identify strengths and weaknesses of learning-based prediction models with regard to e.g., different traffic agent behaviors and traffic scenarios.

The in-depth analysis of datasets should not only be limited to the dynamic properties of agents in the dataset (e.g., velocity distribution), but should also include the distribution of the maneuvers present in the dataset (e.g., left turn maneuver, lane change maneuver).
This requires methods that are able to combine both the vehicle dynamics and the underlying information about the static infrastructure given by an High Definition (HD) map.
The extraction of maneuvers from datasets does not only allow an in-depth analysis, but does also allow training maneuver prediction models.
In this case, the labeled maneuvers serve as the ground-truth.

\begin{figure}[tpb]
	\vspace{0.0cm}
	\centering
	\includegraphics[width=\columnwidth, trim=0cm 0cm 0cm 0cm, clip]{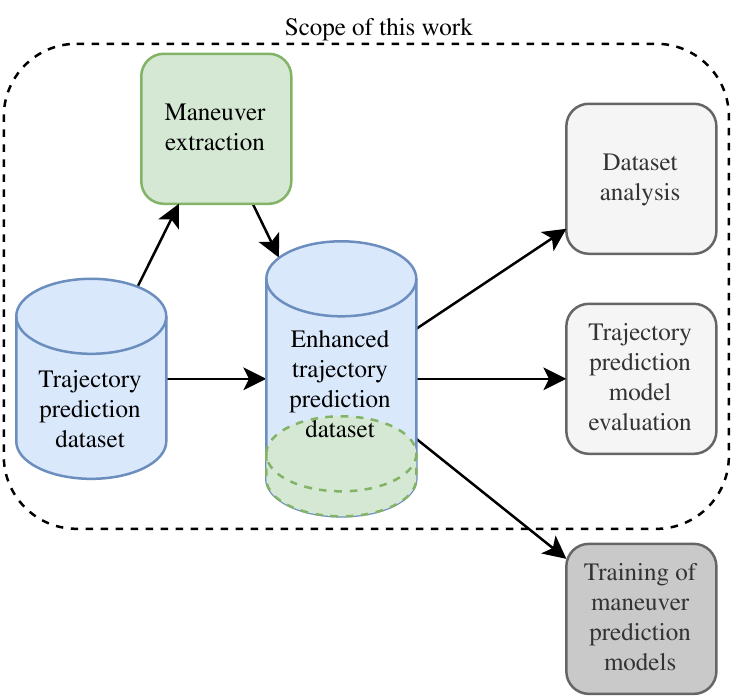}
	\caption{We propose a methodology to extract maneuvers from agent trajectories. As shown in this work, the maneuvers offer a unified way to analyze and compare different trajectory prediction datasets and allow an extensive evaluation of trajectory prediction models. We leave it open to future work to use the methodology in order to train maneuver prediction models.}
	\label{fig:motivation}
	\vspace{-0.37cm}
\end{figure}

With this work we aim to tackle the analysis of trajectory prediction datasets and the evaluation of learning-based trajectory prediction models for different maneuvers.
Maneuvers are extracted under consideration of the agent dynamics and the lane segments the agent traveled along.
An overview of the scope of this work is given in Fig. \ref{fig:motivation}. 
In summary, our main contributions are:
\begin{itemize}
	\item We propose MEAT \textit{(Maneuver Extraction from Agent Trajectories)}, a methodology to extract driven maneuvers from agents trajectories. MEAT is not limited to the dynamic properties of agents, but also considers information provided by the lane graph of HD maps.
	\item We analyze two state-of-the-art trajectory prediction datasets with regard to the dynamics of vehicles and the maneuvers extracted with our methodology.
	\item Based on this analysis, we extensively evaluate multiple state-of-the-art learning-based trajectory prediction models.
\end{itemize}
\section{Related Work}
The related work discussed in this section includes relevant datasets for trajectory prediction, the availability of labeled maneuvers and automated labeling strategies for such datasets, and existing approaches to analyze such datasets.

\subsection{Trajectory Prediction Datasets}
Recent trajectory prediction datasets consist of labeled agent trajectories and static infrastructure given by an HD map.
Due to its lightweight approach, Argoverse Motion Forecasting Dataset 1.1 \cite{Chang2019}, subsequently referred to as Argoverse, is one of the most used datasets.
In Argoverse, an agent trajectory is defined as a timeseries of $x$ and $y$ coordinates in a global coordinate frame.
The HD map is limited to a graph of lane segments, which we now call lane graph, and additional semantic information of the lane segments (e.g., turn direction, has traffic control measure).

Waymo Open Dataset \cite{Sun2020}, subsequently called Waymo, extends this lightweight representation.
Agent trajectories contain not only 3D global coordinates, but also additional state information (e.g., velocity, yaw, bounding box dimensions) and information about the semantic agent class (e.g., car, pedestrian).
In comparison to Argoverse, the HD map is more detailed, not only containing the lane graph, but also traffic signs, traffic lights and crosswalks.

Other publicly available datasets are Lyft Level 5, NuScenes and  Argoverse 2.
Lyft Level 5 \cite{Houston2020} contains information with a similar level of detail as Waymo.
NuScenes \cite{Caesar2020} lacks additional state information of dynamic agents.
While Argoverse 2 \cite{Wilson2021} contains additional state information, both NuScenes and Argoverse 2 contain HD maps that are not as detailed as the one of Waymo and Lyft Level 5.

The datasets listed above are all recorded from vehicles equipped with sensors.
An alternative group of datasets is based on the footage of drones and traffic cameras, allowing for highly accurate recordings.
These datasets include INTERACTION \cite{Zhan2019_ARXIV} and highD \cite{Krajewski2018}.
Due to the stationary recording setups, however, the number of different locations in these datasets is limited.

\subsection{Maneuvers in Agent Trajectories}
Predicting the maneuver an agent located in the surroundings of the autonomous vehicle will execute in its near future is a task closely related to trajectory prediction.
Recent approaches apply deep learning techniques, such as Recurrent Neural Networks \cite{Xu2019}, Convolutional Neural Networks \cite{Mersch2021, Casas2018} or Graph Neural Networks \cite{Li2020a} in order to predict the agent's maneuver.

The amount of publicly available datasets that contain labeled maneuvers is small.
Honda Research Institute Driving Dataset (HDD) \cite{Ramanishka2018} contains recordings with manually labeled maneuvers (e.g., intersection passing, left turn, right turn).
However, the dataset is missing the HD map, which makes it unsuitable for state-of-the-art prediction models.
The highD dataset \cite{Krajewski2018} also contains labeled maneuvers.
Due to the fact that it is a drone-recorded dataset, the scenarios are limited to six different locations only.

In addition to the lack in publicly available datasets with labeled maneuvers, there is a lack in automated labeling strategies for maneuvers that involve the HD map for labeling.
HDD and multiple datasets used in \cite{Mylavarapu2020} for learning-based maneuver prediction are manually annotated and contain no HD map.
Furthermore, the dataset used in \cite{Casas2018} is manually annotated.
highD uses an automated way to extract maneuvers considering lane markings of a map during the labeling process.
However, the labeling process is tailored to highway scenarios.
The crossing of lane markings results in a labeled lane change maneuver.
Taking into account that driving around an obstacle or a noisy trajectory should not result in a lane change label suggests that there is room for improvement.

\subsection{In-Depth Analysis of Trajectory Prediction Datasets}
The proprietary descriptions of the mentioned datasets are limited to high level metadata or statistics.
Because the formats of the descriptions vary from dataset to dataset, no direct comparison is possible.
Glasmacher et al. \cite{Glasmacher2022} propose an automated analysis framework for trajectory datasets.
The framework allows the calculation of interaction, anomaly and relevance scores for agent trajectories.
While this offers a unified methodology to analyze any dataset containing trajectories, the resulting scores fail to provide fundamental and interpretable information such as the information whether the analyzed trajectory reflects a turning process or not.
Our methodology dispenses metascores and allows deriving fundamental and interpretable maneuvers in an automated fashion.
\section{Maneuver Extraction Methodology} \label{sec:maneuver_extraction}
This section describes MEAT, our novel methodology to extract maneuvers from agent trajectories under consideration of the lane graph provided by an HD map.
Our description refers to the extraction of the maneuvers of one agent only.
However, the methodology can be applied to all agents in a scene in parallel.

On a high level, there are two stages:
Firstly, a sequence of lane segments driven by the agent is determined.
For this, the entire trajectory of the agent is considered.
Secondly, driven maneuvers are derived using information about the driven lane segments and their connectivity.
While in this work we only use the methodology to analyze datasets and to evaluate maneuver-specific performances of trajectory prediction models, the methodology can also be used to train and evaluate maneuver prediction models. 

\subsection{Determination of the Driven Lane Sequence}
\begin{figure}[tpb]
	\vspace{0.23cm}
	\centering
	\includegraphics[width=\columnwidth, trim=5.5cm 21.5cm 19.5cm 4.8cm, clip]{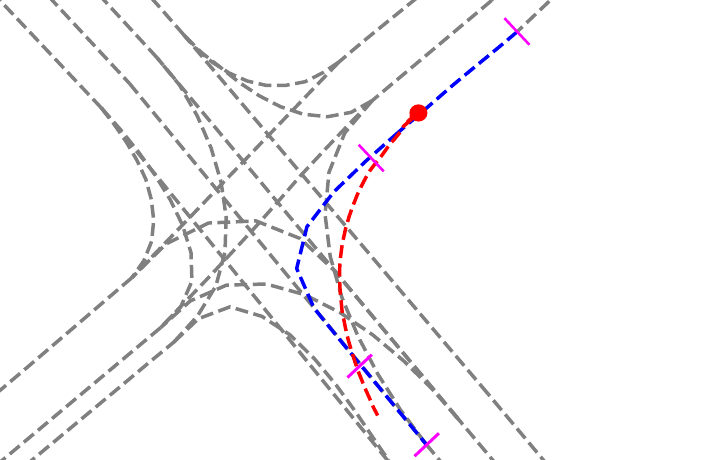}
	\caption{Exemplary result of the lane sequence determination. The driven trajectory of the agent is colored in red, with the starting point being marked with a red dot. The centerline of the extracted lane sequence is colored in blue, with the beginning and end of lane segments being marked with a pink line. Centerlines of other lane segments are colored in grey.}
	\label{fig:example}
\end{figure}

\begin{enumerate}
    \item For each timestep $t$ in the agent's trajectory, an assignment confidence to all lane segments in a fixed radius is calculated. Based on \cite{Petrich2013}, we calculate the assignment confidence by normalizing the shortest Euclidean distance $d^{(t)}_{k}$ between the centerline of a lane segment $k$ and the agent with the threshold distance $d_{\mathrm{th}}$. This can be denoted as $p^{(t)}_{k} =  \max \left( 0, 1 - \frac{d^{(t)}_{k}}{d_{\mathrm{th}}} \right)$.
    Assignments that exceed a predefined confidence threshold $p_{\mathrm{th}}$ result in a probabilistic agent-lane assignment.
    \item Based on the timestep-wise agent-map assignments, all agent-lane assignment intervals are calculated. An agent-lane assignment interval contains a lane segment and the interval of timesteps the agent is assigned to this lane segment. If an agent is assigned to a lane segment in multiple non-contiguous intervals, multiple agent-lane assignment intervals are generated.
    \item Starting from the lane segments to which the agent is assigned to at the first observable timestep, a depth-first search is applied. The depth-first search extracts a set of lane sequences. Each lane sequence in this set contains one valid sequence of lane segments, starting from a lane segment to which the agent is assigned to at the first observable timestep and ending at a lane segment to which the agent is assigned to at the last observable timestep. Valid lane sequences are lane sequences where the consecutive lane segments to which the agent is assigned to all have a valid connectivity, i.e., successor or lane change connections. 
    Connectivity information is provided by the lane graph.
    \item For each resulting lane sequence, a maneuver confidence is calculated. The maneuver confidence is the mean of the timestep-wise agent-map assignment confidences.
\end{enumerate}
For the purpose of our dataset evaluation, we extract only the lane sequence with the highest maneuver confidence.
$d_{\mathrm{th}} = \SI{5}{m}$ and $p_{\mathrm{th}} = 0.5$ are used as the threshold values.
Fig. \ref{fig:example} shows the extracted lane sequence and its lane segments of one exemplary agent trajectory.

\subsection{Derivation of the Resulting Maneuver}
The knowledge about the lane sequence an agent traveled along, allows to derive the following maneuvers:
\begin{itemize}
    \item Assuming that the dataset contains information of a lane segment's turn direction, we can directly derive the turn maneuvers \textbf{going straight}, \textbf{turning left}, \textbf{turning right} and \textbf{both}: If one or multiple of the lane segments in the lane sequence contain the same turn direction, we can directly derive turning left and turning right. If the lane segments of one lane sequence contain different turn directions, we derive the both maneuver. If none of the lane segments contains a turn direction, we derive the straight maneuver. For datasets that do not contain information about a lane segment's direction, this information is automatically generated based on the lane segment curvature, the orientation difference of the first and the last lane segment and the fact that a lane segment in the sequence has two or more predecessors.
    \item Given the lane sequence and the connectivity of the lane segments in this sequence (connectivity is again provided by the lane graph), the lane change maneuvers \textbf{following lane}, \textbf{changing lane left}, \textbf{changing lane right} and \textbf{both} are derivable.
\end{itemize}

The derived maneuvers in this work are limited to those described above.
However, given the lane sequence, it is possible to extract arbitrary complex maneuvers, which even include static infrastructure beyond lanes:
For instance, given a lane segment's turn direction and the information that a crosswalk crosses the lane segment, it is possible to derive a turning into crosswalk maneuver.
We leave the derivation of more complex maneuvers open for future work.
MEAT therefore lifts the main restrictions of prior work by allowing to extract arbitrary complex maneuvers without being limited to specific scenarios.
It has to be noted, however, that similar to \cite{Krajewski2018}, driving around an obstacle or noisy trajectories can still result in unwanted lane change labels.
Temporal filtering of the lane sequence would be one way to prevent this effect.
\section{Analysis of Datasets}
We analyze two state-of-the-art trajectory prediction datasets:
\begin{itemize}
	\item \textbf{Argoverse Motion Forecasting Dataset 1.1} \cite{Chang2019} consists of $201$k sequences in the train and $39$k sequences in the validation split.
	Each sequence is sampled with \SI{10}{Hz} and contains $5$ seconds of vehicle recordings in the cities Miami and Pittsburgh.
	Given the first $2$ seconds of the recording, the goal is to predict the subsequent $3$ seconds.
	Predictions are only evaluated for one selected agent in the sequence.
	For this reason, both our dataset analysis and our analysis of prediction models refer to exactly this agent.
	\item \textbf{Waymo Open Dataset} \cite{Sun2020} consists of $72$k segments in the train and $16$k segments in the validation split.
	Each segment has a duration of $20$ seconds and is sampled with \SI{10}{Hz}.
	Full segments are broken down into $9$ second sequences with a stride of $4$ seconds.
	Given the first second of a sequence, the goal is to predict the subsequent $8$ seconds of up to $8$ selected agents.
	Again, our analyses refer to these agents.
\end{itemize}

The following dataset analysis is divided into two parts.
Firstly, the distributions of dynamic properties of the agent trajectories are analyzed.
Dynamic properties are the average agent velocity, the average agent acceleration and the maximum curvature of the lane segments the agent traveled along.
Secondly, the distribution of maneuvers are analyzed.
The lane sequences required for the calculation of the maximum curvature and the maneuvers are derived by the methodology described in Section \ref{sec:maneuver_extraction}.

\begin{figure*}[thpb]
	\centering
	\subfloat[Argoverse]{%
		\includegraphics[width=0.33\textwidth,trim={0 0 0 0.0cm},clip]{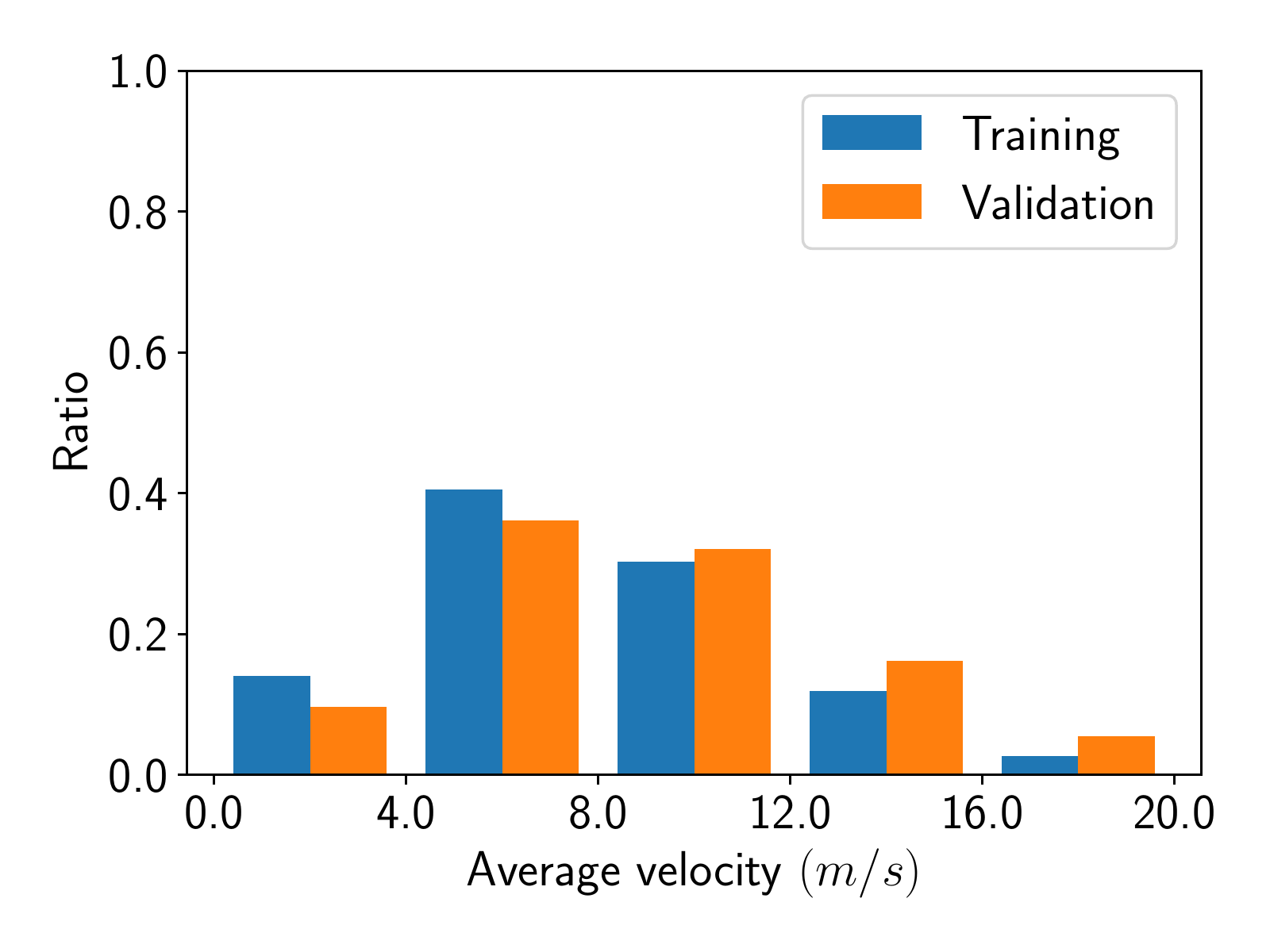}
		\hfill
		\includegraphics[width=0.33\textwidth,trim={0 0 0 0.0cm},clip]{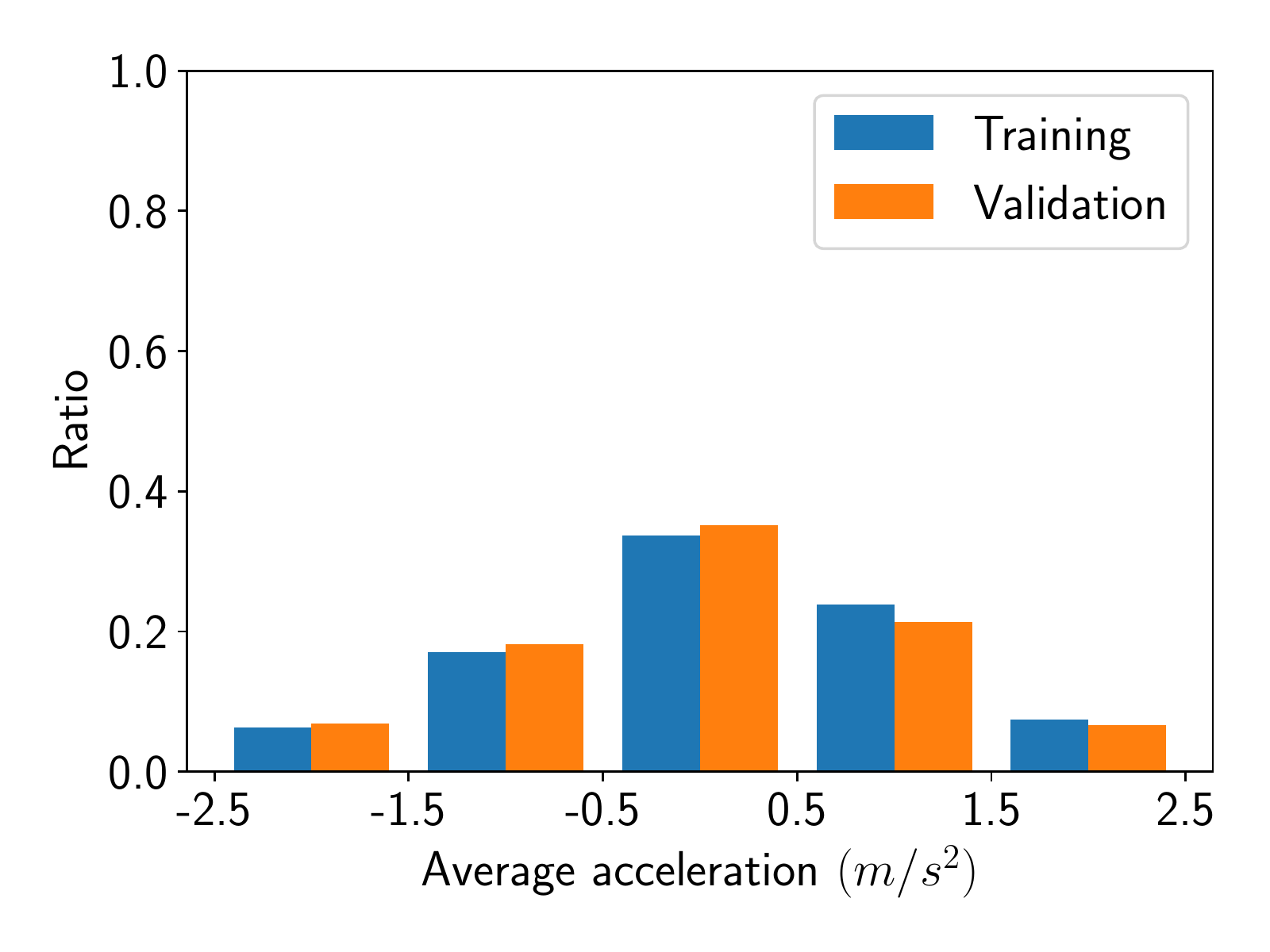}
		\hfill
		\includegraphics[width=0.33\textwidth,trim={0 0 0 0.0cm},clip]{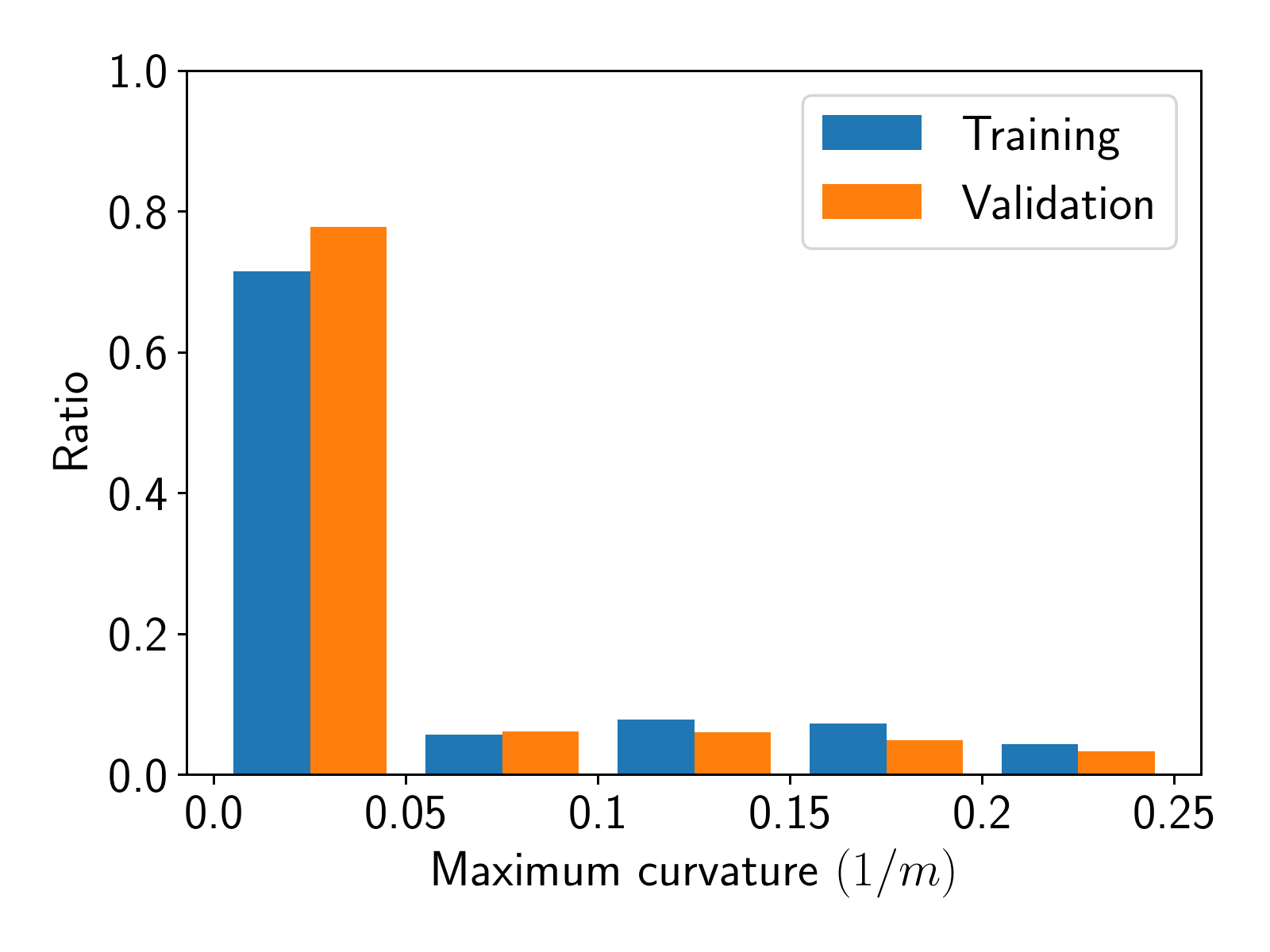}
	}
	\\
	\centering
	\subfloat[Waymo]{%
		\includegraphics[width=0.33\textwidth,trim={0 0 0 0.0cm},clip]{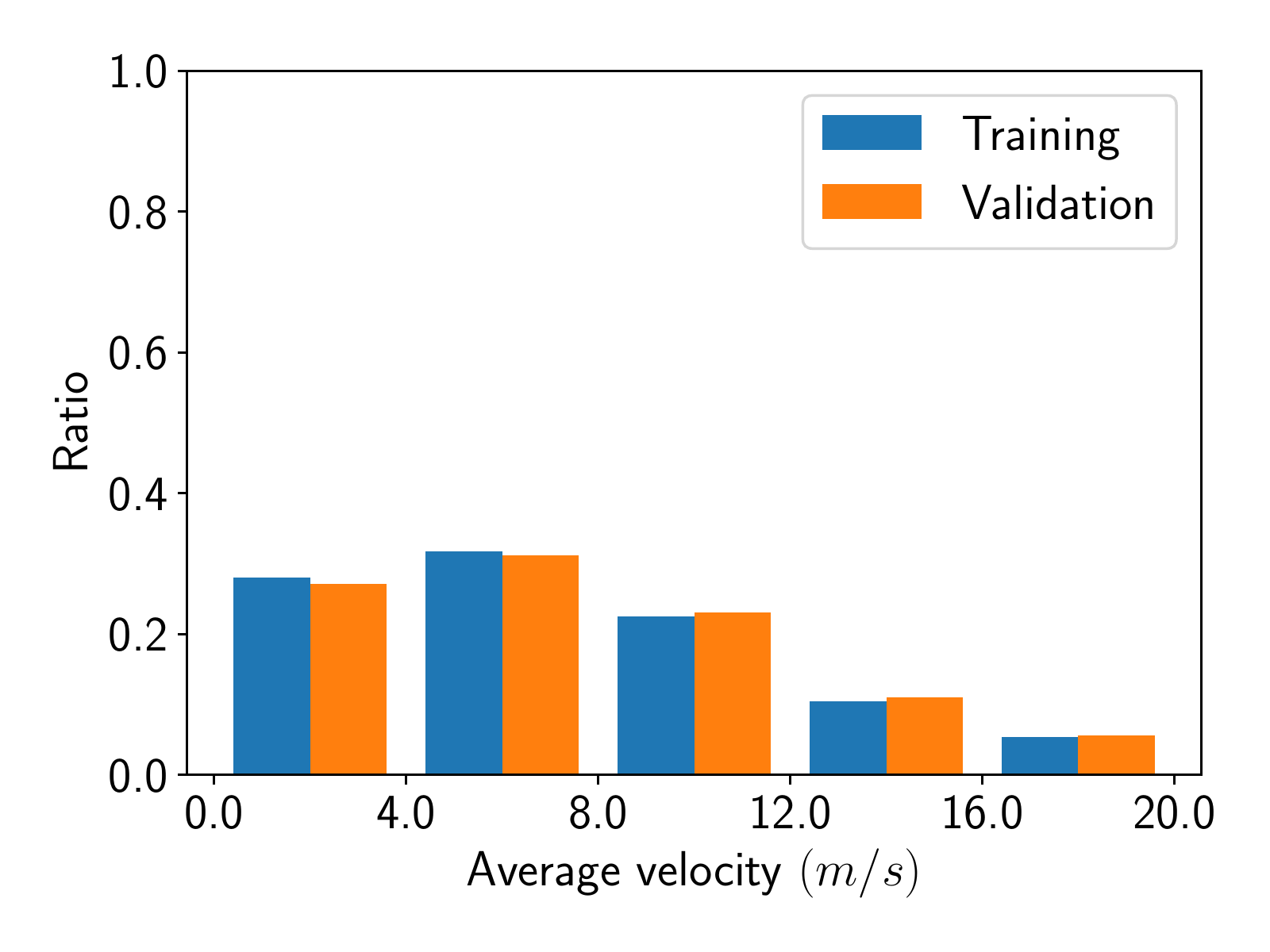}
		\hfill
		\includegraphics[width=0.33\textwidth,trim={0 0 0 0.0cm},clip]{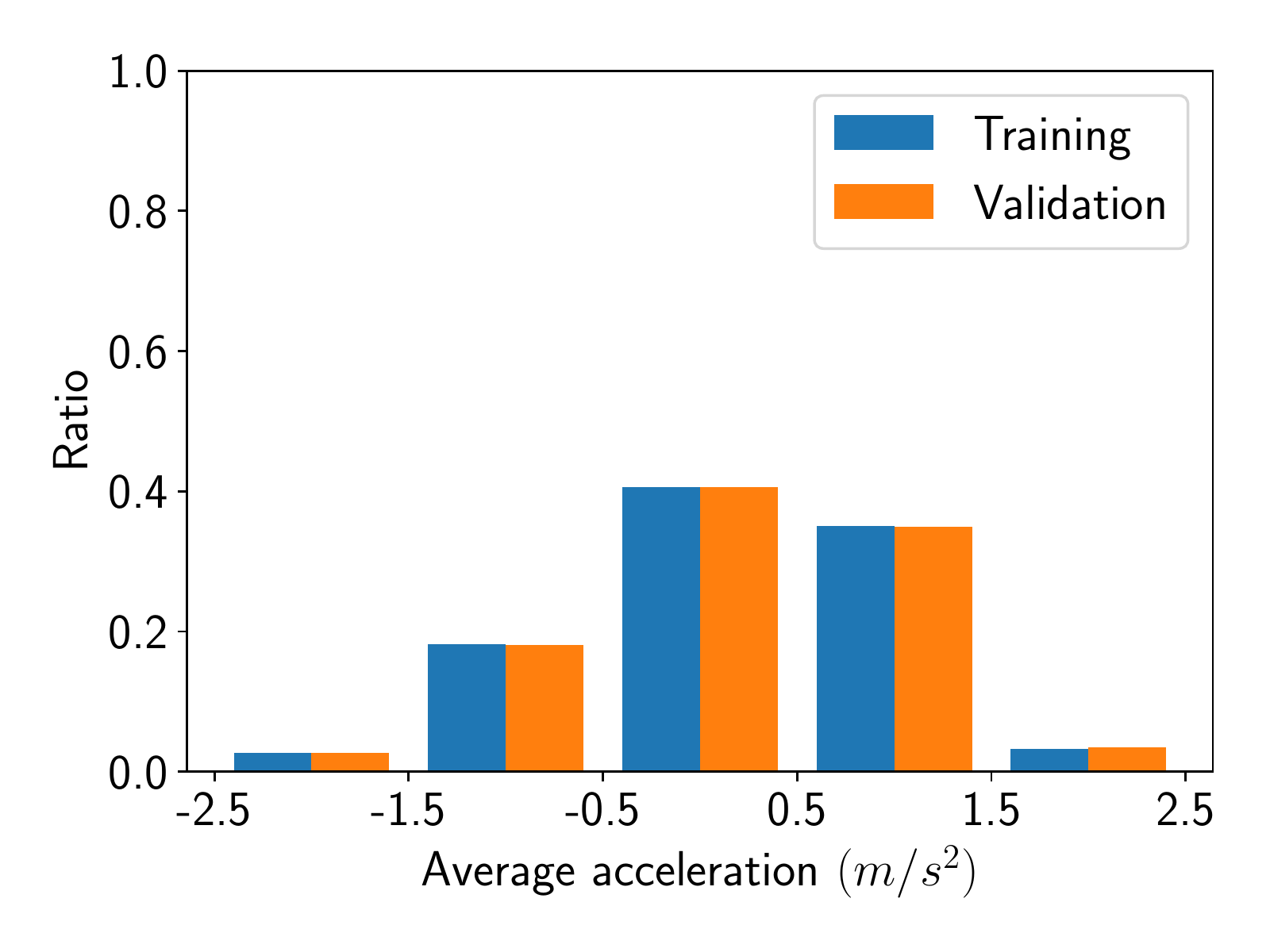}
		\hfill
		\includegraphics[width=0.33\textwidth,trim={0 0 0 0.0cm},clip]{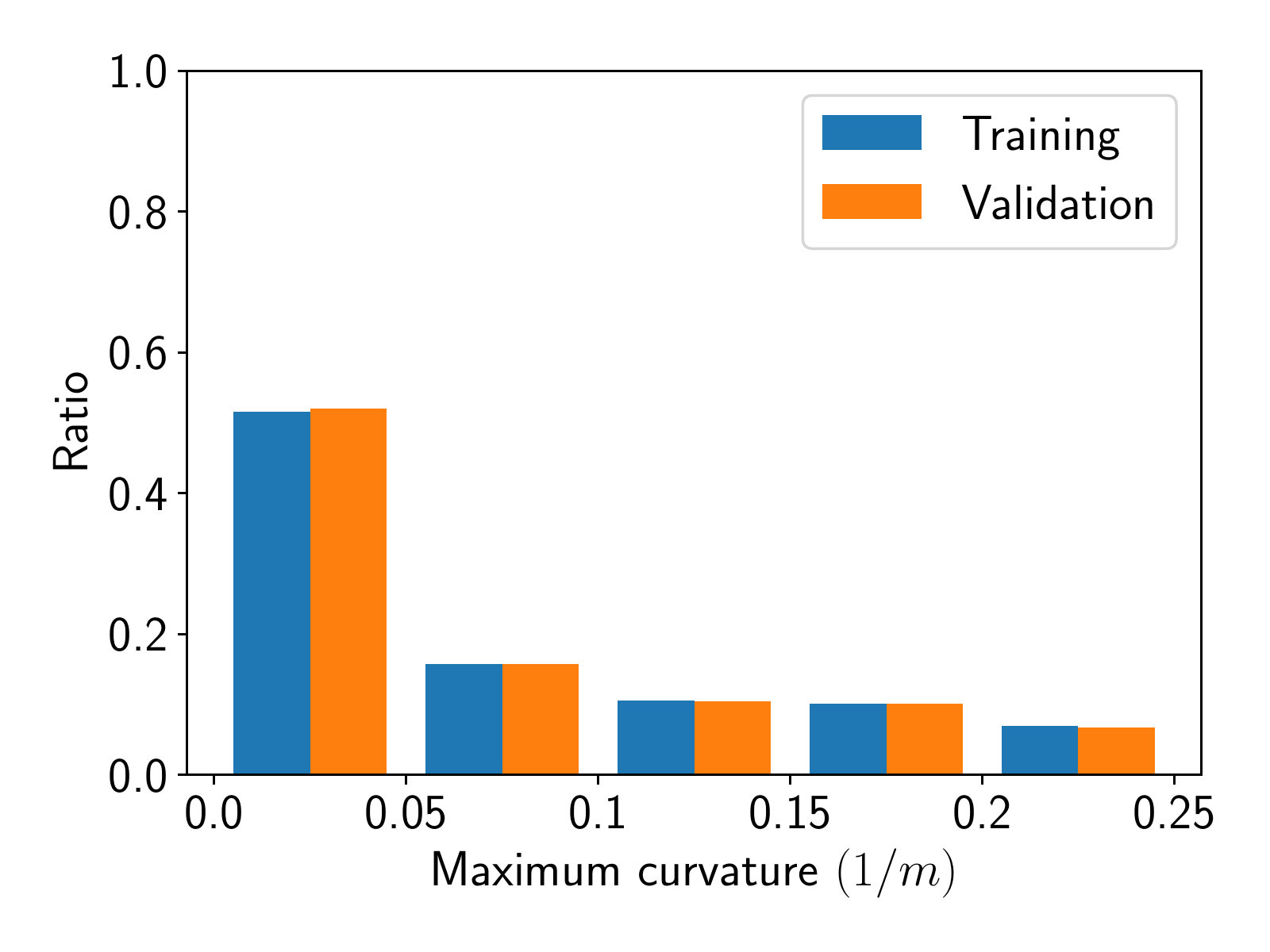}
	}
	\caption{Dataset analysis: Dynamics. Distribution of the average agent velocity (left), the average agent acceleration (center) and the maximum curvature of lane segments the agent traveled along (right) for the train and validation split of Argoverse (a) and Waymo (b).}
	\label{fig:qualitative_results_dynamics}
\end{figure*}

\begin{figure*}[thpb]
	\centering
	\subfloat[Argoverse]{%
		\includegraphics[width=0.33\textwidth,trim={0 0 0 0.0cm},clip]{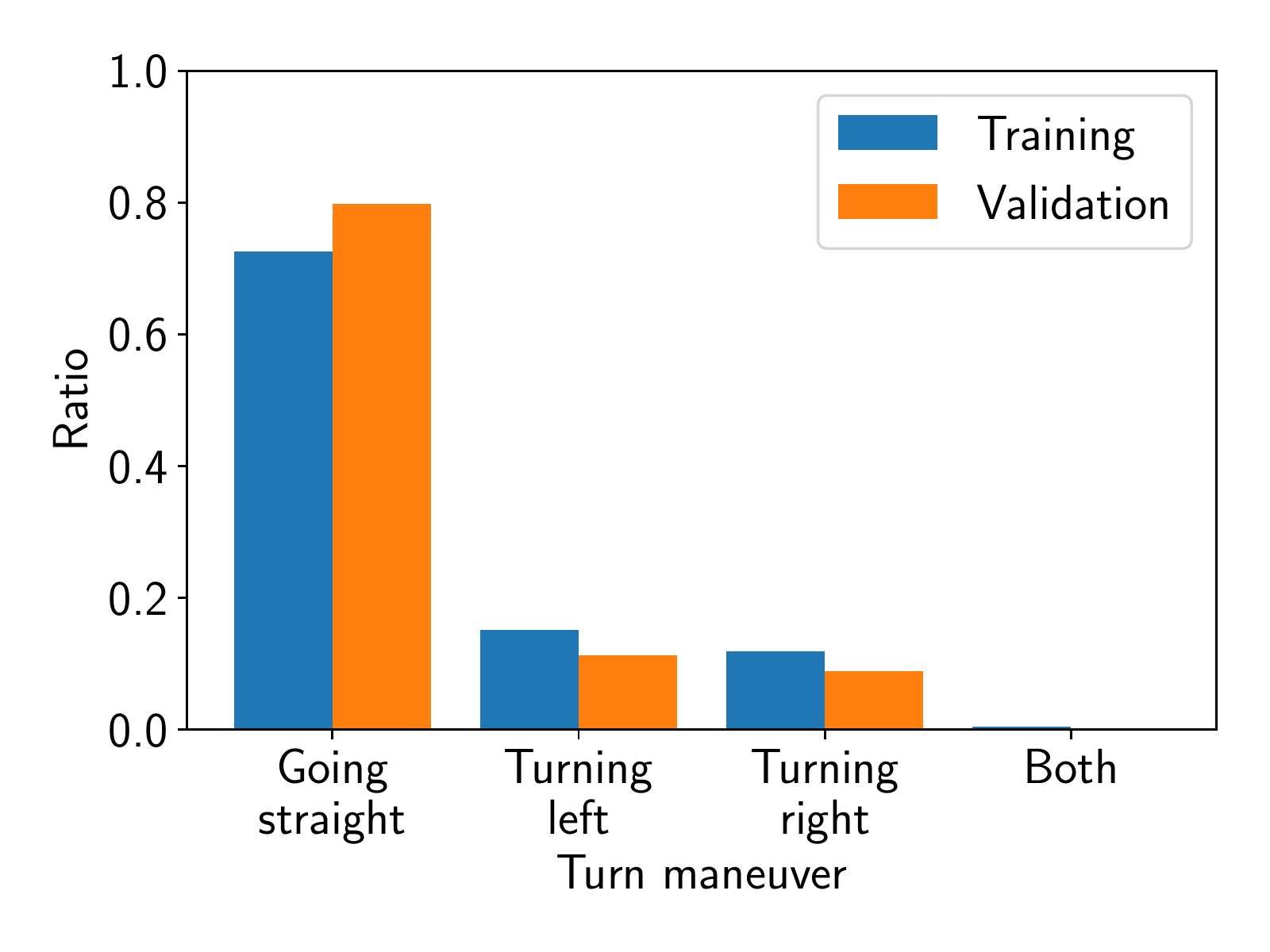}
		\hfill
		\includegraphics[width=0.33\textwidth,trim={0 0 0 0.0cm},clip]{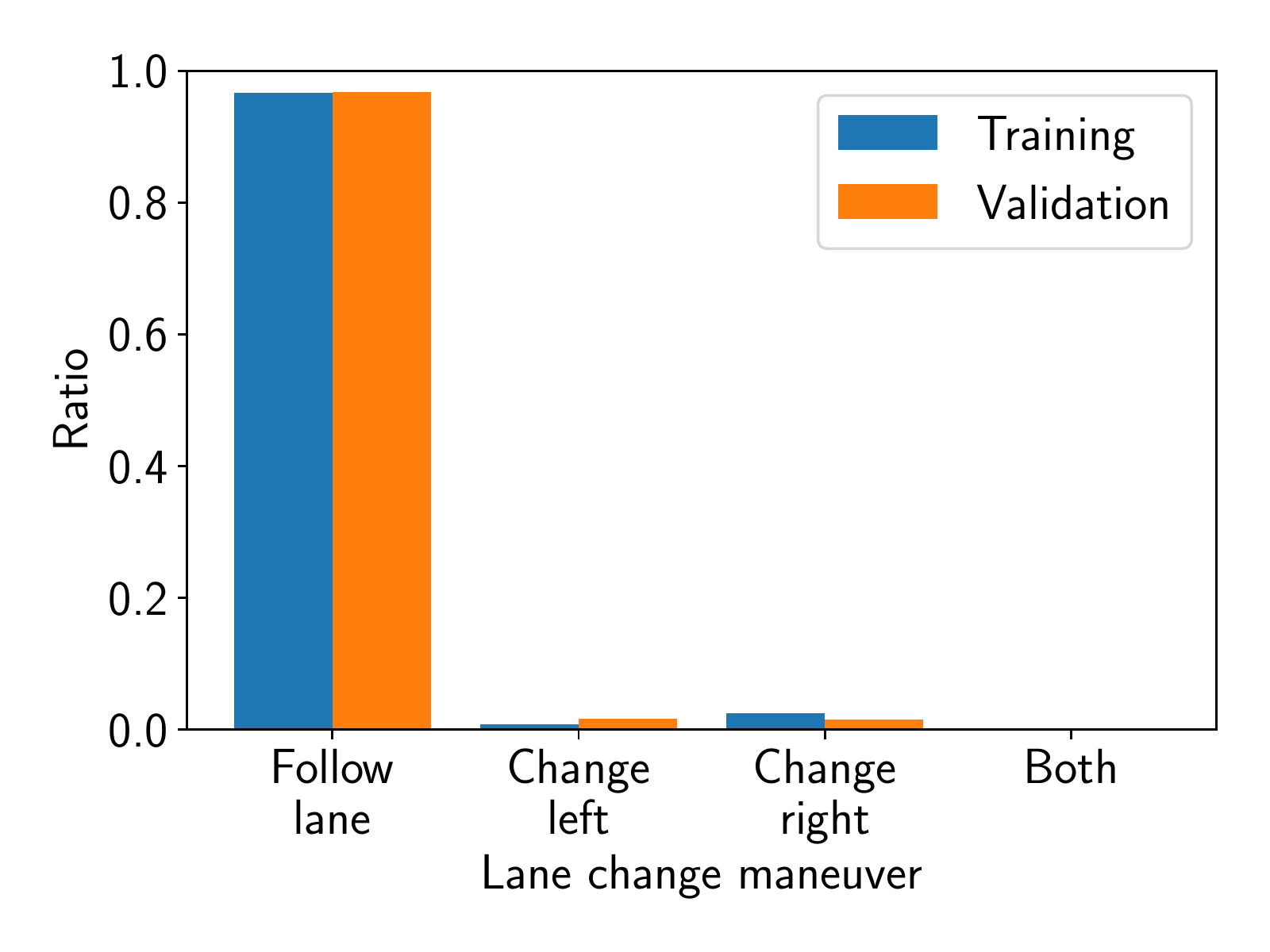}
	}
	\\
	\centering
	\subfloat[Waymo]{%
		\includegraphics[width=0.33\textwidth,trim={0 0 0 0.0cmm},clip]{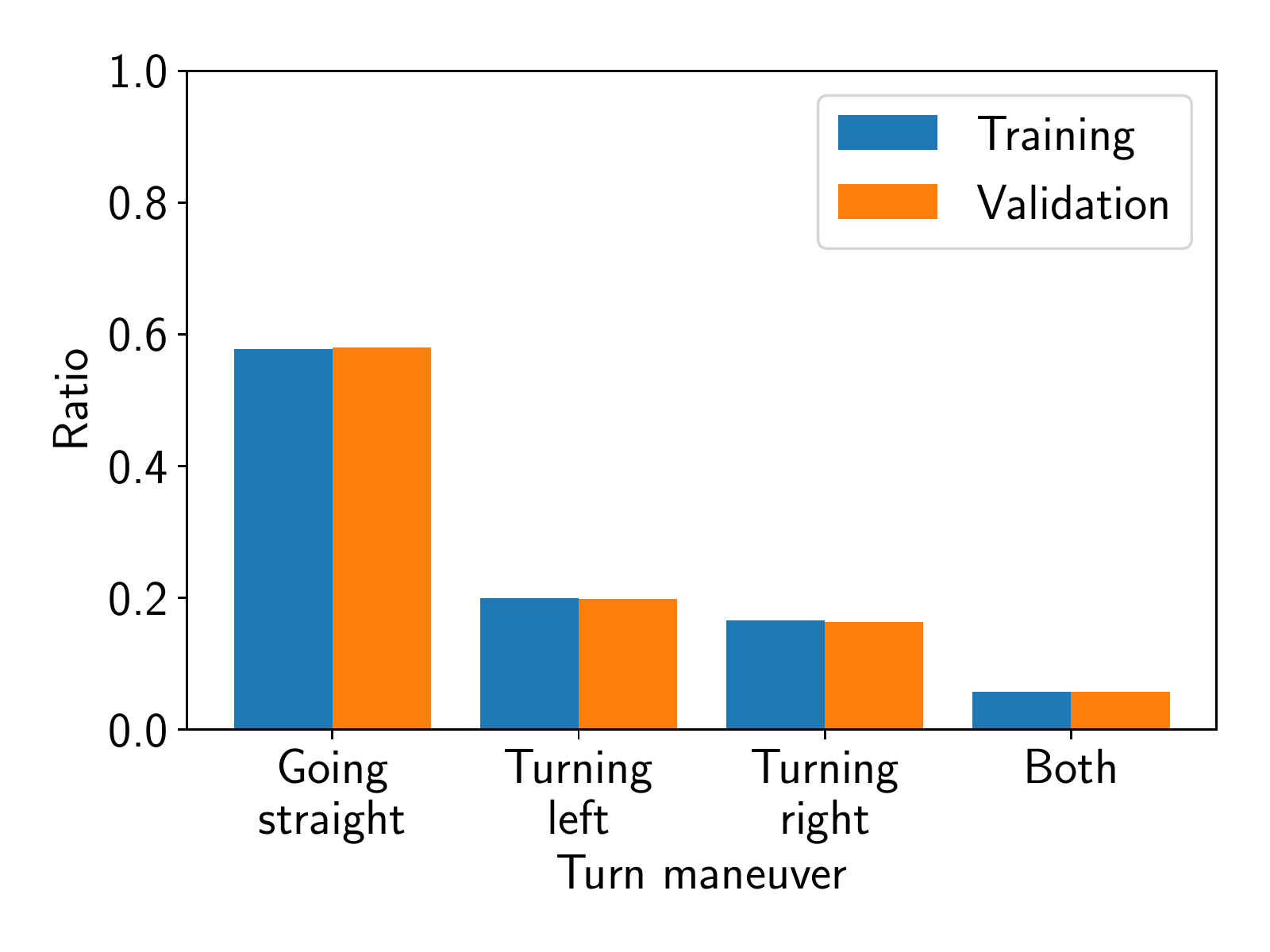}
		\hfill
		\includegraphics[width=0.33\textwidth,trim={0 0 0 0.0cm},clip]{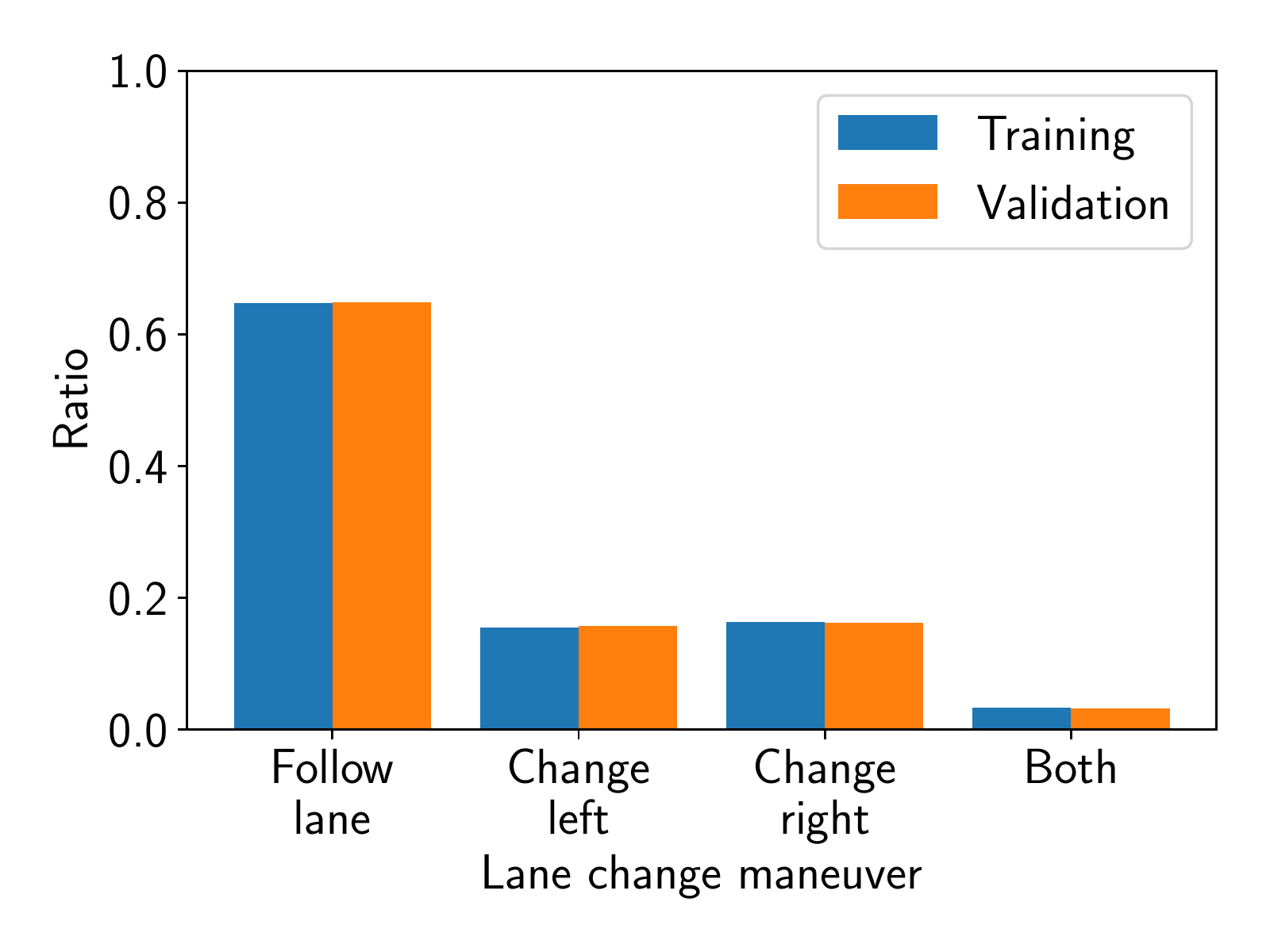}
	}
	\caption{Dataset analysis: Maneuvers. Distribution of turn maneuvers (left) and lane change maneuvers (right) for the train and validation split of Argoverse (a) and Waymo (b).}
	\label{fig:qualitative_results_maneuvers}
\end{figure*}

Full results are shown in Fig. \ref{fig:qualitative_results_dynamics} and Fig. \ref{fig:qualitative_results_maneuvers}.
In the following, we highlight some interesting observations, but leave it to the reader to make more far-reaching conclusions.

The dynamic analysis (Fig. \ref{fig:qualitative_results_dynamics}) shows that Argoverse contains significantly more agents in the \SI[per-mode=symbol]{4}{\meter\per\second} to \SI[per-mode=symbol]{16}{\meter\per\second} range than Waymo.
Waymo, on the other hand, contains more agents with a velocity $<$\SI[per-mode=symbol]{4}{\meter\per\second}.  
A possible reason for this can be taken directly from the distribution of the maximum curvature:
In Waymo, agents typically move along more curved lanes than in Argoverse.
Intuitively, humans drive slower on curvier roads or more conservative speed limits are used in order to restrict the maximum velocity, which together leads to a higher ratio of low-speed samples.

The maneuver analysis (Fig. \ref{fig:qualitative_results_maneuvers}) suggests that Waymo is more diverse in terms of turn and lane change maneuvers.
Argoverse, on the other hand, consists almost exclusively of straight maneuvers and maneuvers that do not contain a lane change.

A general observation that applies to both, the dynamic and maneuver analysis, is that the distributions between Waymo's train and validation split are much more balanced.
\section{Analysis of Trajectory Prediction Models}
Based on the previous dataset analysis, we evaluate the performance of multiple state-of-the-art trajectory prediction models with regard to the agent dynamics and the maneuvers.
All models are able to predict multi-modal trajectories of agents:

\begin{itemize}
	\item \textbf{LaneGCN} \cite{Liang2020} uses graph convolution operators in order to process the lane graph.
	A four-stages fusion network is then applied to capture interaction between lanes and agents.
	\item \textbf{DenseTNT} \cite{Gu2021} uses a VectorNet \cite{Gao2020} encoder to classify over a set of dense goal candidates. Based on the classified goal candidates, trajectory prediction are generated.
	\item \textbf{CRAT-Pred} \cite{Schmidt2022} is a prediction model achieving state-of-the-art performance without requiring map information.
	Interactions are modeled with graph convolution and self-attention.
\end{itemize}
The source code of LaneGCN, DenseTNT and CRAT-Pred for Argoverse is publicly available.
We refer to the original publications for implementation and training details on Argoverse.
Additionally, we use the default prediction horizon of $3$ seconds, while considering $2$ seconds of history.
In order to allow a comparison across Argoverse and Waymo, we adapted the LaneGCN code to work with Waymo.
Training parameters are adopted from the original implementation for Argoverse.
Again, we use the default prediction horizon of $8$ seconds, while considering $1$ second of history.
On both datasets and for all models, the number of predicted modes is set to $6$.

For evaluation, we use standard metrics for multi-modal trajectory prediction:
\begin{itemize}
	\item minimum Average Displacement Error (\textbf{minADE}) is the minimum average Euclidean error between the ground-truth trajectory and all predicted modes.
	\item minimum Final Displacement Error (\textbf{minFDE}) is the minimum Euclidean error between the endpoint of the ground-truth trajectory and all predicted endpoints.
\end{itemize}

Full results of our prediction model evaluation are shown in Table \ref{tab:model_evaluation_dynamics} and Table \ref{tab:model_evaluation_maneuvers}.
All metrics are calculated for the validation set of both dataset and include mean and standard deviation, indicated by the $\pm$ sign.
In the following, we once again highlight some interesting observations, but leave it to the reader to make more far-reaching conclusions.

The dynamic evaluation (Table \ref{tab:model_evaluation_dynamics}) shows the prediction performance of the models for different average velocities, average accelerations and maximum lane curvatures.
While for Waymo the minADE and minFDE consistently increase with the velocity, the picture on Argoverse is not as clear: For the \SI[per-mode=symbol]{8}{\meter\per\second} to \SI[per-mode=symbol]{12}{\meter\per\second} range, the minADE is lower than for the neighboring velocity ranges and the minFDE is the lowest of all velocity ranges across all models.
For acceleration, the models show a lower minFDE for braking maneuvers (\SI[per-mode=symbol]{-2.5}{\meter\per\second\squared} to \SI[per-mode=symbol]{-1.5}{\meter\per\second\squared}) than for accelerating maneuvers (\SI[per-mode=symbol]{1.5}{\meter\per\second\squared} to \SI[per-mode=symbol]{2.5}{\meter\per\second\squared}). 

Comparing the map-free CRAT-Pred model to the map-aware LaneGCN and DenseTNT models shows that the minFDE of the map-free model for the velocity range \SI[per-mode=symbol]{16}{\meter\per\second} to \SI[per-mode=symbol]{20}{\meter\per\second} is lower than for velocities $<$\SI[per-mode=symbol]{8}{\meter\per\second}.
This is not the case for the map-aware models.

The maneuver evaluation (Table \ref{tab:model_evaluation_maneuvers}) confirms the intuitive assumption that more complex maneuvers result in higher prediction errors.
Both datasets are recorded in right-hand drive regions, meaning that right turn maneuvers have a lower level of complexity than left turn maneuvers.
However, the errors resulting from right turn maneuvers are higher than the errors resulting from left turn maneuvers (Argoverse) or very similar to each other (Waymo).
We assume that this is because both datasets have a higher distribution of left turn maneuvers than right turn maneuvers in the training set.
The prediction performance related to lane change maneuvers reflects an opposite trend:
Right lane change maneuvers are predicted with lower errors than left lane change maneuvers.
A possible reason for this is that right lane changes are usually triggered by a predictable reason, e.g., pulling onto an exit ramp.
Left lane changes, on the other hand, are more often triggered by more arbitrary events, e.g., overtaking a leading vehicle.

\begin{table*}[thpb]
	\vspace{0.23cm}
	\caption{Evaluation of trajectory prediction models: Dynamics}
	\label{tab:model_evaluation_dynamics}
	\scriptsize
	\setlength{\tabcolsep}{2.3pt}
	\centering
	\setlength\extrarowheight{-1pt}
	\begin{tabularx}{1\textwidth}{Xlllllllllllllllll}
		\toprule
		& \multirow{2}{*}{Model}     & \multirow{2}{*}{Metric} &                       \multicolumn{5}{c}{Average velocity (\si{\metre\per\second})}                        &                   \multicolumn{5}{c}{Average acceleration (\si{\metre\per\second\squared})}                    &            \multicolumn{5}{c}{Maximum curvature (\SI{1e-2}{\per\metre})}             \\
		\cmidrule(lr){4-8}\cmidrule(lr){9-13}\cmidrule(lr){14-18}             &                            &                         & \tiny$[0, 4)$   & \tiny$[4, 8)$   & \tiny$[8, 12)$  & \tiny$[12, 16)$ & \tiny$[16, 20]$ & \tiny$[-2.5, -1.5)$ & \tiny$[-1.5, -0.5)$ & \tiny$[-0.5, 0.5)$ & \tiny$[0.5, 1.5)$ & \tiny$[1.5, 2.5]$ & \tiny$[0, 5)$   & \tiny$[5, 10)$  & \tiny$[10, 15)$ & \tiny$[15, 20)$ & \tiny$[20, 25]$ \\ \midrule \midrule
		\parbox[t]{2mm}{\multirow{12}{*}{\rotatebox[origin=c]{90}{Argoverse}}} & \multirow{4}{*}{LaneGCN}   & \multirow{2}{*}{minADE}   & $0.61$& $0.70$& $0.66$& $0.74$& $0.80$& $0.79$& $0.67$& $0.62$& $0.67$& $0.77$& $0.64$& $1.04$& $0.86$& $0.85$& $0.85$\\ 
		& && \tiny$\pm 0.44$& \tiny$\pm 0.55$& \tiny$\pm 0.57$& \tiny$\pm 0.78$& \tiny$\pm 0.85$& \tiny$\pm 0.66$& \tiny$\pm 0.63$& \tiny$\pm 0.52$& \tiny$\pm 0.59$& \tiny$\pm 0.67$& \tiny$\pm 0.57$& \tiny$\pm 1.04$& \tiny$\pm 0.72$& \tiny$\pm 0.57$& \tiny$\pm 0.61$\\  \cmidrule(lr){3-18} 
		& & \multirow{2}{*}{minFDE}   & $1.02$& $1.09$& $0.95$& $1.07$& $1.12$& $1.09$& $1.02$& $0.92$& $1.05$& $1.19$& $0.88$& $1.91$& $1.45$& $1.47$& $1.58$\\ 
		& && \tiny$\pm 0.94$& \tiny$\pm 1.13$& \tiny$\pm 1.21$& \tiny$\pm 1.59$& \tiny$\pm 1.63$& \tiny$\pm 1.34$& \tiny$\pm 1.35$& \tiny$\pm 1.00$& \tiny$\pm 1.12$& \tiny$\pm 1.41$& \tiny$\pm 1.00$& \tiny$\pm 2.50$& \tiny$\pm 1.61$& \tiny$\pm 1.26$& \tiny$\pm 1.48$\\  \cmidrule(lr){2-18} 
		& \multirow{4}{*}{DenseTNT}   & \multirow{2}{*}{minADE}   & $0.67$& $0.72$& $0.70$& $0.80$& $0.92$& $0.82$& $0.72$& $0.66$& $0.71$& $0.83$& $0.68$& $1.13$& $0.92$& $0.89$& $0.84$\\ 
		& && \tiny$\pm 0.54$& \tiny$\pm 0.62$& \tiny$\pm 0.67$& \tiny$\pm 0.95$& \tiny$\pm 1.10$& \tiny$\pm 0.74$& \tiny$\pm 0.77$& \tiny$\pm 0.67$& \tiny$\pm 0.73$& \tiny$\pm 0.82$& \tiny$\pm 0.72$& \tiny$\pm 1.29$& \tiny$\pm 0.85$& \tiny$\pm 0.78$& \tiny$\pm 0.60$\\  \cmidrule(lr){3-18} 
		& & \multirow{2}{*}{minFDE}   & $1.00$& $1.06$& $0.94$& $1.02$& $1.15$& $1.08$& $1.01$& $0.89$& $1.02$& $1.24$& $0.87$& $1.95$& $1.42$& $1.48$& $1.40$\\ 
		& && \tiny$\pm 1.12$& \tiny$\pm 1.32$& \tiny$\pm 1.51$& \tiny$\pm 1.97$& \tiny$\pm 2.29$& \tiny$\pm 1.60$& \tiny$\pm 1.69$& \tiny$\pm 1.29$& \tiny$\pm 1.47$& \tiny$\pm 1.92$& \tiny$\pm 1.39$& \tiny$\pm 3.15$& \tiny$\pm 1.96$& \tiny$\pm 1.95$& \tiny$\pm 1.31$\\  \cmidrule(lr){2-18} 
		& \multirow{4}{*}{CRAT-Pred}   & \multirow{2}{*}{minADE}   & $0.80$& $0.87$& $0.78$& $0.87$& $0.94$& $0.92$& $0.80$& $0.75$& $0.84$& $0.93$& $0.75$& $1.31$& $1.14$& $1.19$& $1.11$\\ 
		& && \tiny$\pm 0.63$& \tiny$\pm 0.78$& \tiny$\pm 0.74$& \tiny$\pm 0.95$& \tiny$\pm 0.91$& \tiny$\pm 0.77$& \tiny$\pm 0.79$& \tiny$\pm 0.68$& \tiny$\pm 0.81$& \tiny$\pm 0.86$& \tiny$\pm 0.68$& \tiny$\pm 1.31$& \tiny$\pm 1.09$& \tiny$\pm 1.02$& \tiny$\pm 1.01$\\  \cmidrule(lr){3-18} 
		& & \multirow{2}{*}{minFDE}   & $1.48$& $1.57$& $1.25$& $1.39$& $1.46$& $1.49$& $1.35$& $1.25$& $1.49$& $1.66$& $1.16$& $2.67$& $2.22$& $2.42$& $2.25$\\ 
		& && \tiny$\pm 1.52$& \tiny$\pm 1.86$& \tiny$\pm 1.76$& \tiny$\pm 2.15$& \tiny$\pm 1.93$& \tiny$\pm 1.78$& \tiny$\pm 1.83$& \tiny$\pm 1.52$& \tiny$\pm 1.83$& \tiny$\pm 2.09$& \tiny$\pm 1.39$& \tiny$\pm 3.26$& \tiny$\pm 2.80$& \tiny$\pm 2.66$& \tiny$\pm 2.56$\\
		\cmidrule(lr){1-18}
		\parbox[t]{2mm}{\multirow{4}{*}{\rotatebox[origin=c]{90}{Waymo}}} & \multirow{4}{*}{LaneGCN}   & \multirow{2}{*}{minADE}   & $0.32$& $0.51$& $0.61$& $0.73$& $0.78$& $0.75$& $0.57$& $0.52$& $0.47$& $0.69$& $0.54$& $0.53$& $0.51$& $0.51$& $0.52$\\ 
		& && \tiny$\pm 0.37$& \tiny$\pm 0.63$& \tiny$\pm 0.92$& \tiny$\pm 1.41$& \tiny$\pm 1.75$& \tiny$\pm 1.19$& \tiny$\pm 0.90$& \tiny$\pm 1.00$& \tiny$\pm 0.76$& \tiny$\pm 1.05$& \tiny$\pm 1.02$& \tiny$\pm 0.87$& \tiny$\pm 0.79$& \tiny$\pm 0.78$& \tiny$\pm 0.81$\\  \cmidrule(lr){3-18} 
		& & \multirow{2}{*}{minFDE}   & $0.49$& $0.83$& $0.94$& $1.02$& $1.02$& $0.83$& $0.68$& $0.76$& $0.84$& $1.40$& $0.73$& $0.89$& $0.88$& $0.87$& $0.90$\\ 
		& && \tiny$\pm 0.64$& \tiny$\pm 0.94$& \tiny$\pm 1.20$& \tiny$\pm 1.68$& \tiny$\pm 1.99$& \tiny$\pm 1.25$& \tiny$\pm 0.94$& \tiny$\pm 1.20$& \tiny$\pm 1.15$& \tiny$\pm 1.85$& \tiny$\pm 1.24$& \tiny$\pm 1.18$& \tiny$\pm 1.10$& \tiny$\pm 1.07$& \tiny$\pm 1.13$\\
		\bottomrule                                                           
	\end{tabularx}
\end{table*}

\begin{table*}[thpb]
	\caption{Evaluation of trajectory prediction models: Maneuvers}
	\label{tab:model_evaluation_maneuvers}
	\scriptsize
	\setlength{\tabcolsep}{10pt}
	\centering
	\setlength\extrarowheight{-1pt}
	\begin{tabularx}{1\textwidth}{Xllllllllllll}
		\toprule
		& \multirow{2}{*}{Model}     & \multirow{2}{*}{Metric} &                     \multicolumn{4}{c}{Turn maneuver}                     &                 \multicolumn{4}{c}{Lane change maneuver}                  \\
		\cmidrule(lr){4-7}\cmidrule(lr){8-11}                              &                            &                         & \tiny Going straight  & \tiny Turning left      & \tiny Turning right     & \tiny Both      & \tiny Following lane    & \tiny Changing lane left      & \tiny Changing lane right     & \tiny Both              \\ \midrule\midrule
		\parbox[t]{2mm}{\multirow{12}{*}{\rotatebox[origin=c]{90}{Argoverse}}}                  & \multirow{4}{*}{LaneGCN}   & \multirow{2}{*}{minADE}   & $0.65$& $0.83$& $0.90$& $1.07$& $0.69$& $1.03$& $0.88$& $1.16$\\ 
		& && \tiny$\pm 0.61$& \tiny$\pm 0.67$& \tiny$\pm 0.69$& \tiny$\pm 0.58$& \tiny$\pm 0.62$& \tiny$\pm 0.95$& \tiny$\pm 0.57$& \tiny$\pm 0.65$\\  \cmidrule(lr){3-11} 
		& & \multirow{2}{*}{minFDE}   & $0.92$& $1.42$& $1.60$& $1.46$& $1.02$& $1.83$& $1.50$& $1.90$\\ 
		& && \tiny$\pm 1.18$& \tiny$\pm 1.38$& \tiny$\pm 1.58$& \tiny$\pm 1.21$& \tiny$\pm 1.24$& \tiny$\pm 2.09$& \tiny$\pm 1.34$& \tiny$\pm 1.20$\\  \cmidrule(lr){2-11} 
		& \multirow{4}{*}{DenseTNT}   & \multirow{2}{*}{minADE}   & $0.71$& $0.87$& $0.91$& $0.98$& $0.73$& $1.12$& $0.90$& $1.29$\\ 
		& && \tiny$\pm 0.78$& \tiny$\pm 0.77$& \tiny$\pm 0.78$& \tiny$\pm 0.49$& \tiny$\pm 0.78$& \tiny$\pm 1.09$& \tiny$\pm 0.86$& \tiny$\pm 0.41$\\  \cmidrule(lr){3-11} 
		& & \multirow{2}{*}{minFDE}   & $0.93$& $1.35$& $1.51$& $1.35$& $1.01$& $1.82$& $1.50$& $2.09$\\ 
		& && \tiny$\pm 1.61$& \tiny$\pm 1.69$& \tiny$\pm 1.85$& \tiny$\pm 0.95$& \tiny$\pm 1.63$& \tiny$\pm 2.34$& \tiny$\pm 2.07$& \tiny$\pm 0.66$\\  \cmidrule(lr){2-11} 
		& \multirow{4}{*}{CRAT-Pred}   & \multirow{2}{*}{minADE}   & $0.77$& $1.08$& $1.22$& $1.89$& $0.83$& $1.16$& $1.13$& $1.28$\\ 
		& && \tiny$\pm 0.72$& \tiny$\pm 1.01$& \tiny$\pm 1.10$& \tiny$\pm 2.36$& \tiny$\pm 0.81$& \tiny$\pm 1.06$& \tiny$\pm 1.00$& \tiny$\pm 0.66$\\  \cmidrule(lr){3-11} 
		& & \multirow{2}{*}{minFDE}   & $1.21$& $2.09$& $2.51$& $3.52$& $1.41$& $2.17$& $2.09$& $2.26$\\ 
		& && \tiny$\pm 1.55$& \tiny$\pm 2.46$& \tiny$\pm 2.86$& \tiny$\pm 5.57$& \tiny$\pm 1.86$& \tiny$\pm 2.45$& \tiny$\pm 2.52$& \tiny$\pm 1.22$\\ 
		\cmidrule(lr){1-11}
		\parbox[t]{2mm}{\multirow{4}{*}{\rotatebox[origin=c]{90}{Waymo}}} & \multirow{4}{*}{LaneGCN}   & \multirow{2}{*}{minADE}   & $0.52$& $0.55$& $0.53$& $0.54$& $0.50$& $0.60$& $0.54$& $0.57$\\ 
		& && \tiny$\pm 0.98$& \tiny$\pm 0.90$& \tiny$\pm 0.81$& \tiny$\pm 0.66$& \tiny$\pm 0.93$& \tiny$\pm 1.00$& \tiny$\pm 0.84$& \tiny$\pm 0.77$\\  \cmidrule(lr){3-13} 
		& & \multirow{2}{*}{minFDE}   & $0.71$& $0.93$& $0.89$& $1.01$& $0.70$& $1.01$& $0.95$& $1.05$\\ 
		& && \tiny$\pm 1.20$& \tiny$\pm 1.21$& \tiny$\pm 1.11$& \tiny$\pm 1.06$& \tiny$\pm 1.14$& \tiny$\pm 1.34$& \tiny$\pm 1.14$& \tiny$\pm 1.24$\\
		\bottomrule
	\end{tabularx}
\end{table*}
\section{Conclusion}
In this work we propose an automated way to extract maneuvers from agent trajectories.
Based on the extracted maneuvers, our in-depth dataset analysis and the extensive evaluation of prediction models give valuable insight into dataset balance and maneuver-specific prediction performance.
We leave it open to future work to investigate the usability of our extracted maneuvers for the training of learning-based maneuver prediction models.
Furthermore, it remains to be investigated whether the simultaneous learning of maneuver and trajectory prediction leads to synergy effects.

\addtolength{\textheight}{-15.3cm}

\section*{Acknowledgment}
The research leading to these results is funded by the German Federal Ministry for Economic Affairs and Climate Action within the project "KI Delta Learning" (F\"orderkennzeichen 19A19013A). The authors would like to thank the consortium for the successful cooperation.

\bibliographystyle{IEEEtran}
\bibliography{Literature}

\end{document}